\documentclass[journal]{IEEEtranTIE}
\usepackage{graphicx}
\usepackage{cite}
\usepackage{picinpar}
\usepackage{amsmath}
\usepackage{url}
\usepackage{flushend}
\usepackage[latin1]{inputenc}
\usepackage{colortbl}
\usepackage{soul}
\usepackage{multirow}
\usepackage{pifont}
\usepackage{color}
\usepackage{alltt}
\usepackage[hidelinks]{hyperref}
\usepackage{enumerate}
\usepackage{siunitx}
\usepackage{breakurl}
\usepackage{epstopdf}
\usepackage{pbox}
\usepackage[vlined,ruled]{algorithm2e}
\usepackage{caption}
\usepackage{subcaption}
\usepackage{booktabs} 
\usepackage{multirow}  
\usepackage{xcolor}
\usepackage{threeparttable}
\usepackage{graphics}
\usepackage{amssymb}
\usepackage{booktabs}

\begin{document}

\title{ Anti-Degeneracy Scheme for Lidar SLAM based on Particle Filter in Geometry Feature-Less Environments }

\author
{
	\vskip 1em
	
    Yanbin Li, Wei Zhang*, \emph{IEEE Member}, Zhiguo Zhang, Xiaogang Shi, Ziruo Li, \\Mingming Zhang,  Hongping Xie and Wenzheng Chi, \emph{IEEE Senior Member}

	\thanks
    {
		This work is supported by National Science Foundation of China grant \#62273246 and by Science and Technology Research
        Foundation of State Grid Co.Ltd (5700-202318270A-1-1-ZN).
		(* corresponding author)

		Yanbin Li, Wei Zhang, Zhiguo Zhang, Xiaogang Shi are with School of Electronics Engineering, Beijing University of Posts and Telecommunications, Beijing 100876, China
        (yanbinli@bupt.edu.cn; weizhang13@bupt.edu.cn; zhangzhiguo@bupt.edu.cn;
        shixiaogang@bupt.edu.cn).
        Ziruo Li is with Key Lab of Smart Agriculture Systems, Ministry of Education, China Agricultural University, Beijing 100083, China (liziruo123@cau.edu.cn).
        Mingming Zhang is with School of Integrated Circuit Science and Engineering, Beihang University, Beijing 100191, China (zmm@buaa.edu.cn)
        Hongping Xie is with Co.Ltd Construction Branch, State Grid Jiangsu Electric Power, Jiangsu 226000, China (sg\_supporting@126.com).
        Wenzheng Chi is with the Robotics and Microsystems Center, School of Mechanical and Electric Engineering, Soochow University, Suzhou 215021, China (wzchi@suda.edu.cn).
	}
}

\maketitle

\begin{abstract}
Simultaneous localization and mapping (SLAM) based on particle filtering has been extensively employed in indoor scenarios due to its high efficiency. 
However, in geometry feature-less scenes, the accuracy is severely reduced due to lack of constraints. 
In this article, we propose an anti-degeneracy system based on deep learning. 
Firstly, we design a scale-invariant linear mapping to convert coordinates in continuous space into discrete indexes, in which a data augmentation method based on Gaussian model is proposed to ensure the model performance by effectively mitigating the impact of changes in the number of particles on the feature distribution.
Secondly, we develop a degeneracy detection model (DD-Model) using residual neural networks (ResNet) and transformer which is able to identify degeneracy by scrutinizing the distribution of the particle population.
Thirdly, an adaptive anti-degeneracy strategy is designed, which first performs fusion and perturbation on the resample process to provide rich and accurate initial values for the pose optimization, and use a hierarchical pose optimization combining coarse and fine matching, which is able to adaptively adjust the optimization frequency and the sensor trustworthiness according to the degree of degeneracy, in order to enhance the ability of searching the global optimal pose.
Finally, we demonstrate the optimality of model, as well as the improvement of image matrix method and GPU on the inference time through ablation experiments. The effectiveness of our method is verified by comparing with SOTA methods. 
\end{abstract}



\definecolor{limegreen}{rgb}{0.2, 0.8, 0.2}
\definecolor{forestgreen}{rgb}{0.13, 0.55, 0.13}
\definecolor{greenhtml}{rgb}{0.0, 0.5, 0.0}

\begin{figure}[!t]\centering
	\includegraphics[width=9cm]{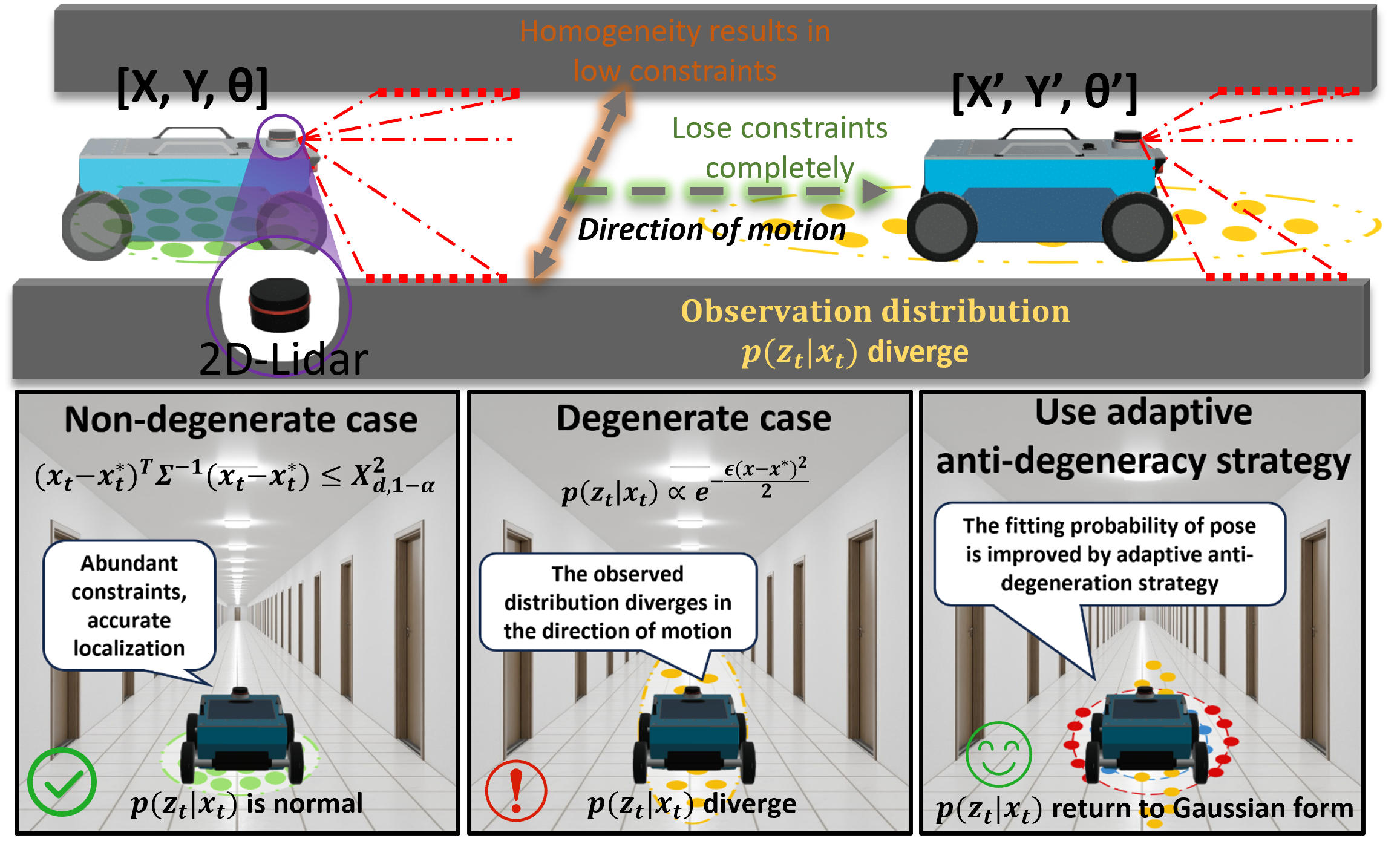}
    \vspace{-18pt}
	\caption{Anti-degeneracy strategy restores observation distribution to an approximate Gaussian distribution, improving the search for the optimal pose.}\label{problem_archi_show}
    \vspace{-15pt}
\end{figure}

\section{Introduction}


\IEEEPARstart{S}{lam} is the core of realizing navigation for robots, which enables devices to perform real-time localization and map construction in unknown environments.
2D-SLAM is widely used in the localization task of service robots and vacuum cleaners because of its low cost and high reliability \cite{grisetti2007improved,7487258}, and its output two-dimensional grid map is suitable for path planning and obstacle avoidance.

However, in geometry feature-less scenarios, such as long straight corridors, underground pipelines and spacious halls, the accuracy of SLAM is relatively low. 
This is due to the sparseness of the environmental features causes SLAM to lack sufficient constraints for accurate self-localization.
In monotonous color walls or scenes with repetitive textures, camera-based SLAM systems have difficulty extracting enough feature points for tracking, resulting in degeneracy of positioning accuracy and map quality. 
In addition, in low-light or strongly reflective conditions, the quality of the image captured by the camera can also be significantly degenerate, affecting feature extraction and depth estimation \cite{10753045,10753049,10874215,10423794,10740921,10305271,campos2021orb}.
Lidar-based SLAM is difficult to perform effective feature matching in degenerate environments due to insufficient captured point cloud features. 
Especially in long straight corridors, the lack of orthogonal orientation constraints makes it difficult for the robot to perform accurate localization \cite{zhang2014loam,shan2018lego,shan2020lio,wang2021f}.

Shi $\textit{et al.}$ \cite{shi2022dense} proposed a degeneracy detection method for 2D-SLAM, he achieved detection by extracting normal features from Lidar point clouds and using principal component analysis for feature aggregation and analysis. 
However, this method needs to adjust parameters in different scenarios. The challenge of degeneracy detection is to improve the generalization ability in different scenarios and get rid of the dependence on environmental conditions and preassumptions.

Particle swarm distribution is the most intuitive representation of SLAM convergence state, as shown in Fig. \ref{problem_archi_show}, this distribution pattern can be captured by the neural network to achieve highly generalized degeneracy detection. 
In this article, we propose a detection method for particle-filtering based 2D-SLAM, and design a self-tuning optimization strategy based on the degree of degeneracy. Compared to other robust SLAM methods with fixed optimization strategies like \cite{9981107}, our method can adaptively adjust the optimization strategy based on the level of degeneracy. This flexibility allows the system to better adapt to different environmental conditions. First, it uses linear mapping and Gaussian augmentation to obtain training samples of degeneracy particles. Through supervised learning, a neural network is endowed with degeneracy detection capabilities. Subsequently, the framework employs the model to detect degeneracy by analyzing particle distribution. Finally, we use an optimization framework based on the degree of degeneracy, which adaptively adjusts strategy to enhance the global optimality of pose estimation and change the contributions of different sensors.
The main contributions of this article are summarized as follows:

\begin{enumerate}[1)]
	\item A linear mapping method from continuous space to discrete indexes with scale invariance and a data augmentation strategy based on Gaussian model can ensure the accuracy of particle feature distribution and improve the model performance;
	\item A degeneracy detection model based on ResNet and Transformer architectures, with excellent degeneracy detection capability; and
    \item An adaptive anti-degeneracy strategy that enriches initial values through fusion and perturbation during resampling and employs hierarchical pose optimization with coarse and fine matching. This method adaptively adjusts optimization frequency and sensor trustworthiness based on degeneracy levels, improving the ability to find the global optimal pose.
	
\end{enumerate}


\section{Related work}

\subsection{Degeneracy Detection}

Degeneracy detection is able to identify the degeneracy of sensor in real time, so the system can immediately deploy anti-degeneracy strategies to improve the accuracy using the information about degeneracy.
Lee $\textit{et al.}$ \cite{lee2024switch} detected degeneracy by checking the convergence of the optimization process based on predefined thresholds derived from physical assumptions and statistical significance. Tuna $\textit{et al.}$ \cite{tuna2023x} conducted degeneracy detection through fine-grained localizability analysis based on point cloud correspondences, and determines the degeneracy directions by analyzing the principal components of the optimization directions. Zhang $\textit{et al.}$ \cite{zhang2016degeneracy} achieved degeneracy detection by analyzing the geometric structure of the constraints in the optimization problem and identifying the degeneracy directions through the eigenvalues and eigenvectors of the Hessian matrix. Zhou $\textit{et al.}$ \cite{zhou2020lidar}  estimated the sensor state by minimizing the sum of Mahalanobis norms of all measurement residuals. Chen $\textit{et al.}$ \cite{10816047} proposed a P2d degeneracy detection algorithm that uses adaptive voxel segmentation to integrate local geometric features and calculates degeneracy factors from point cloud distribution changes between frames. However, these methods may perform well in specific environments, yet be less robust when faced with conditions such as rapidly changing environments, and relies on a pre-set threshold.

\subsection{Anti-Degeneracy Strategies}

Anti-degeneracy strategy is the key element to improve SLAM accuracy in degenerate environments.
Li $\textit{et al.}$ \cite{li2022intensity} dealt with the degeneracy problem through geometric and intensity-based feature extraction by designing two multi-weight functions to fully extract the features of planar and edge points. Zhang $\textit{et al.}$ \cite{zhang2024lvio} presented an innovative tightly coupled lidar-vision-inertial odometry, which enables accurate state estimation in degenerate environments through sensor fusion strategies. Zhang $\textit{et al.}$ \cite{zhang2023graph} utilized ARTag as a visual marker to assist in localization, which employs a bitmap optimization method to reduce the error. Lee $\textit{et al.}$ \cite{lee2024switch} proposed a switching-based SLAM that achieves high accuracy by switching from lidar to visual odometry upon detection of lidar degeneracy. 
However, these methods have requirements on the geometry of environment and the strength of features. Meanwhile, the multi-sensor fusion strategy increases the deployment cost, the parameters need to be adjusted for adaptation, which leads to the fact that these methods do not have good robustness.


\section{System overview}

\begin{figure}[!t]\centering
	\includegraphics[width=9cm]{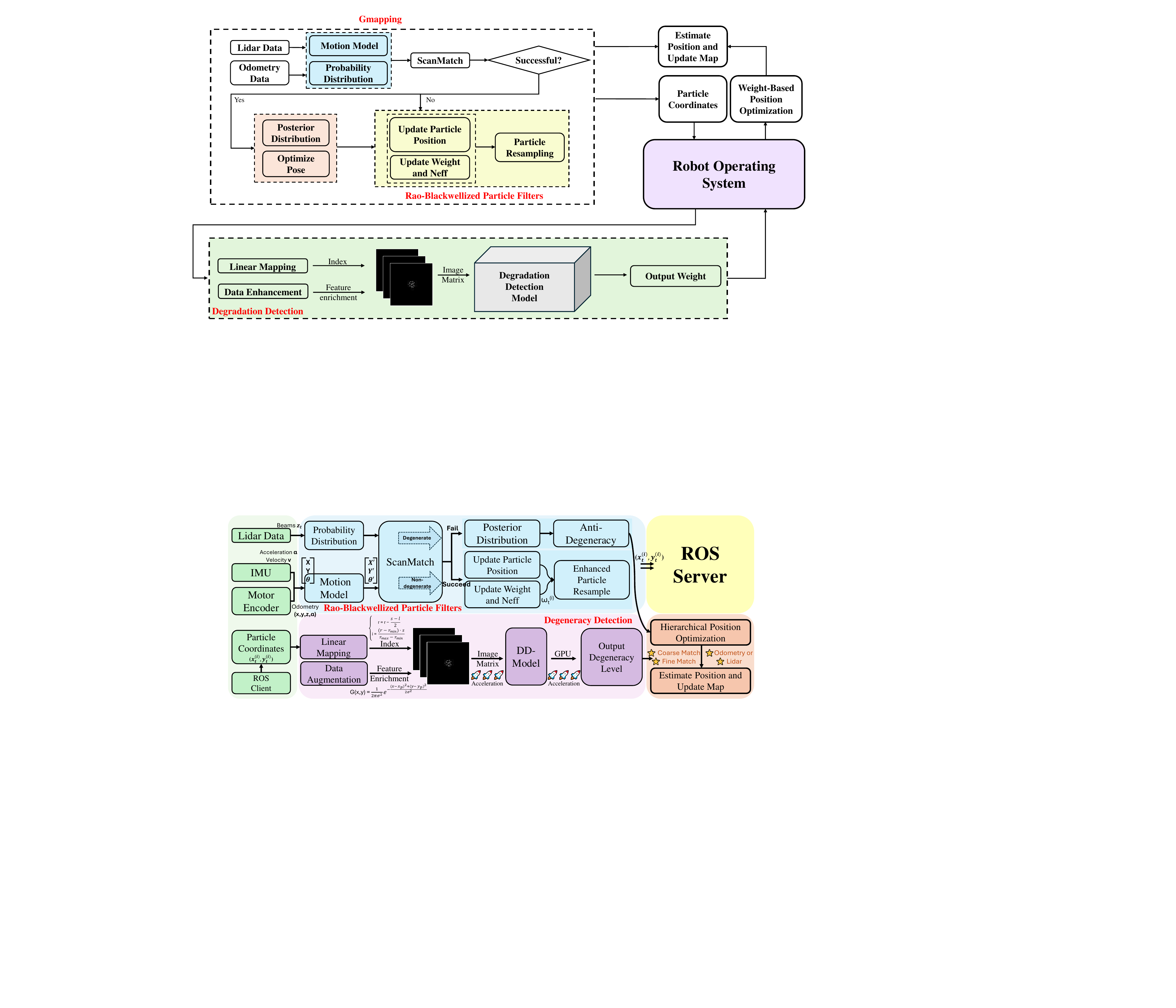}
    \vspace{-18pt}
	\caption{Overall flow chart. Anti-degeneracy strategy will be triggered when scanmatch fails due to degeneracy, increasing the hierarchical pose optimization count by one for every 0.25 increase in degeneracy.}\label{system_pipeline}
    \vspace{-15pt}
\end{figure}

The framework diagram of the anti-degeneracy system presented in this article is shown in Fig. \ref{system_pipeline}. 
Particle filter-based SLAM represents a posteriori probability distribution of robot positions and maps by maintaining a set of random particles \cite{grisetti2007improved}.
Particles are first preliminarily updated using the odometry data from the IMU and motor encoder through the motion model, as shown in (\ref{motion_model}). Subsequently, the observation model is employed to update the particles based on the lidar data to determine the optimal pose through scanmatch of pose estimation. If the environment features are rich enough for scanmatch to succeed, the particle weights which reflect the uncertainty of pose estimation are then calculated according to (\ref{weight}). Finally, resampling and map updating operations are carried out.
\begin{equation}
\label{motion_model}
\begin{bmatrix} X ^ { \prime }  \\ Y ^ { \prime }  \\ \theta ^ { \prime }  \end{bmatrix} = \begin{bmatrix} X \\ Y \\\theta\end{bmatrix} + \begin{bmatrix} \delta _ { t } . \cos ( \theta + \delta _ { r1 } ) \\ \delta _ { t } . \sin ( \theta + \delta _ { r1 } ) \\ \delta _ { r1 } + \delta _ { r2 } \end{bmatrix}
\end{equation}
where $(X, Y) \in \mathbb{R}^2$ represent the position of robot, and $\theta \in \mathbb{S}^1$ represents its orientation. $\delta_t$ is the translation amount, while $\delta_{r1}$ and $\delta_{r2}$ are the rotation amounts before and after translation, respectively.

\begin{equation}
\label{weight}
w _ { i } = w _ { i } \times \prod _ { j = 1 } ^ { n } p ( z _ { j } \vert x _ { i } )
\end{equation}
where $w_i \in \mathbb{R}^+$ and $x_i \in \mathbb{R}^2$ represent the weight and pose of the $i_{th}$ particle, respectively, while $z_j$ denotes the $j_{th}$ measurement and $n$ is the amount of measurement.

As shown in Fig. \ref{problem_archi_show}, in non-degenerate case, SLAM has sufficient environmental constraints. 
The second order taylor expansion is performed on observation likelihood $p(z_t\vert x_t)$ in (\ref{taylor}), where $I_o$ is observation fisher matrix, $x^*$ is groudtruth of pose, $z_t$ and $x_t$ are lidar beam and pose at time $t$ respectively:
\begin{equation}
\label{taylor}
 \log p(z_{t}\vert x_{t})\approx \log p(z_{t}\vert x_{t}^{*})-\frac{1}{2}(x_{t}-x_{t}^{*})^TI_o(x_{t}-x_{t}^{*})
\end{equation}
At this moment, particle distribution radius is constrained by the Mahalanobis distance shown in (\ref{Maanobis}), where $\Sigma=I_o^{-1}$ is posterior covariance matrix , $\chi_{d,1-\alpha}^{2}$ is the critical value of the Chi-square distribution when the degree of freedom is $d$ and the significance level is $a$:
\begin{equation}
\label{Maanobis}
(x_{t}-x_{t}^{*})^{T}\Sigma ^{-1}(x_{t}-x_{t}^{*})\leq \chi_{d,1-\alpha}^{2}
\end{equation}
In this case, all dimensions of covariance are small, and particles cluster around $x^*$. Each particle has a high fitting probability for groundtruth. The dense distribution indicates that the sampling space of pose optimization is relatively certain, which can capture the nonlinear characteristics of the state space and improve the localization accuracy.

In degenerate case, the robot lacks environmental constraints in the travel direction. Particles obtain similar likelihood scores via scanmatch in this direction, causing scanmatch to fail and failing to constrain particle updates. The likelihood distribution $p(z_t\vert x_t)$ diverges in the travel direction and degenerates from gaussian distribution to flat distribution shown in (\ref{likelihood_degenerates}), where $\epsilon$ is the eigenvalue of $I_o$ in the direction of travel:
\begin{equation}
\label{likelihood_degenerates}
p(z_t\vert x_t)\propto exp(-\frac{1}{2}\epsilon(x-x^{*})^{2})
\end{equation}
Then particles cover a wider state space to avoid falling into local optimality and alleviate localization stagnation to some extent. Conversely, in the direction perpendicular to travel, the homogenization of lidar ranging information is severe. Although the constraint information is inaccurate, the severity of the degeneracy problem in the travel direction dominates the divergence direction of particles. This results in compression of the distribution in the perpendicular direction, reducing the fitting probability to the ground truth (GT). The error $\Delta x$ in the direction of travel is projected into the vertical direction through the turning angle $\theta$, causing the map to skew:
\begin{equation}
\label{map_skew}
\Delta y \approx \Delta x \cdot sin \theta
\end{equation}

We transform the degeneracy detection into a classification task for a neural network, analyzing the degeneracy through the dispersion of particles.
We utilize robot operating system (ROS) to transmit particle coordinates to the DD-Model, where they are converted into images via linear mapping and processed using data augmentation based on gaussian, and then the model predicts the degree of degeneracy. If the robot is in a degenerate environment, then scanmatch fails, as shown in Fig. \ref{system_pipeline}, the anti-degeneracy node will improve the pose optimization by hierarchical anti-degeneracy strategies according to the degeneracy level, the likelihood distribution is approximately converted to gaussian form to increase the search capability for the globally optimal pose, and the contribution of different sensors is adjusted by influencing the likelihood score.



\begin{figure*}[t]\centering
	\includegraphics[width=16.7cm]{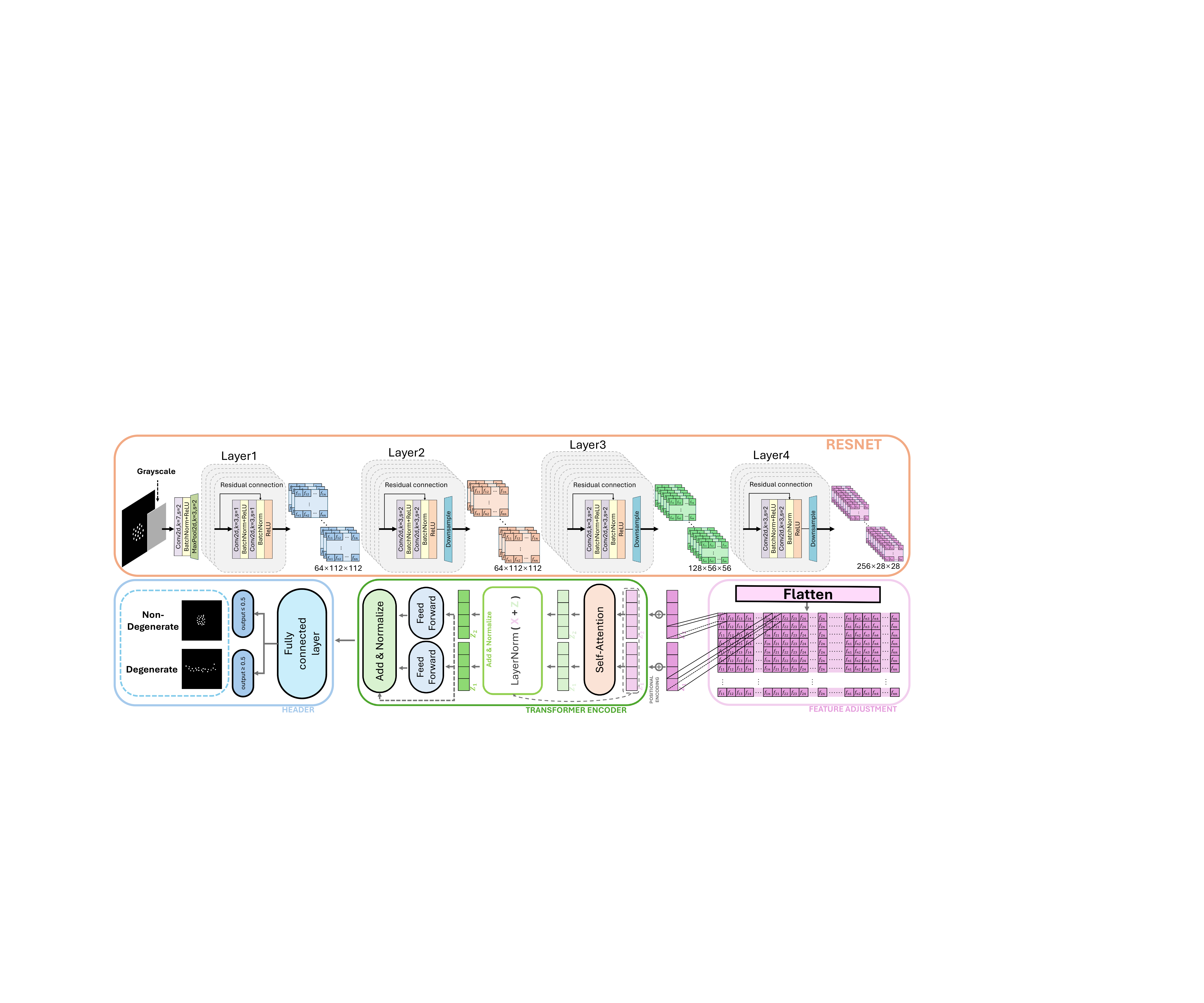}
    \vspace{-5pt}
	\caption{The architecture of DD-Model.}\label{model}
    \vspace{-17pt}
\end{figure*}

\section{METHODOLOGY}

\subsection{Neural Network Based Degeneracy Detection}
\subsubsection{Linear Mapping with Scale Invariance}

The quality of the particle image dataset is crucial for model training. Uncertainty in particle coordinates can distort their relative positions and distribution. We design a mapping method from continuous space to discrete indices for particle swarms of various sizes, ensuring they maintain their original distribution and produce scale-invariant images.

We first define the canvas size of image as 10m in reality, which defines the accuracy threshold for localization and the outliers are considered invalid particles. The increasing number of particles that exceed the boundary leads to a progressively sparser distribution, thereby augmenting the probability that the model will classify the situation as degenerate, consistent with the anticipated outcome.
Subsequently, the image boundary are modified according to (\ref{resize}). The procedure is to first modify the extreme values of the particle coordinates in the x and y dimensions, ensuring that the coordinates in each dimension have the same discrete mapping metric respectively.
\begin{equation}
\label{resize}
\begin{cases} 
r_{\text{min}} = r_{\text{min}} - \frac{s - l}{2}, & \text{if } l_r < s \\ \\
r_{\text{max}} = r_{\text{max}} + \frac{s - l}{2}, & \text{if } l_r < s \\
\end{cases}
\end{equation}
where $r_{max}$ and $r_{min}$ represent the maximum and minimum values of the particle coordinates in each dimension, $l$ is the absolute value of the difference between $r_{max}$ and $r_{min}$, and $s$ is the size of the boundary of the image.

Then we discretize the continuous spatial coordinates of particles into pixel indices, translating physical locations to image space pixels. 
By linearly mapping the coordinate extremes changes in (\ref{resize})  to all particles, the scale invariance can be matained across both the physical and image spaces, as detailed in (\ref{index}), where $i$ is discrete index corresponding to particle coordinates, indicating the pixel position in image.
\begin{equation}
\label{index}
\text{i} =  \frac{(\text{r} - r_{\text{min}}) \cdot \text{s}}{r_{\text{max}} - r_{\text{min}}}, 
\end{equation}

\subsubsection{Data Augmentation based on Gaussian Model}

Particle distributions can be influenced by changes in their number, impacting model performance. For instance, a clustered particle swarm may become sparse if the number decreases. Conversely, too many particles can cause overlap, altering the original distribution.
We adopt the Gaussian model in (\ref{Gaussian}) to simulate the influence of each particle on the surrounding grid, where the position of the particle serves as the center of the Gaussian distribution, and the influence of the particle decreases exponentially with distance.
\begin{equation}
\label{Gaussian}
G(x, y) = \frac{1}{2\pi\sigma^2} \exp\left(-\frac{(x - x_p)^2 + (y - y_p)^2}{2\sigma^2}\right),
\end{equation}
where $(x, y) \in \mathbb{Z}^2$ and $(x_p, y_p) \in \mathbb{Z}^2$ represent the grid positions influenced by the particles and of the particles themselves, respectively, and $\sigma$ is the standard deviation. We adjust it to make the Gaussian influence range a 5*5 grid, as this size optimizes augmentation effects for the typical number of particles and other size range will reduce the effect.

For each grid, we accumulate the contribution of particles to update the occupancy $o$ of this grid:
\begin{equation}
\text{o}(x, y) = \sum_{i=1}^{5}\sum_{j=1}^{5} G(x, y),
\end{equation}
If $o$ exceeds a preset threshold, the grid is marked as occupied. Since it should mark more grids as occupied when overlap is severe, we prefer to set a lower threshold. Notably, when the number of particles is too low, the method will only mark grids in dense particle areas as occupied, while sparse areas remain unchanged, achieving the correct augmentation effect.


\subsubsection{Degeneracy Detection Model}


The overall architecture of the model is shown in Fig. \ref{model}. We first grayscale the image to preserve the contour and shape of the particle swarm and remove redundant information.
Subsequently, the data passes through a convolutional layer with a 7*7 kernel, which facilitates the extraction of features.
Then it passes through a batch normalization (BN) layer with ReLU, which solves bias problem of internal covariates by regularization and enhance the ability of nonlinear representation. 
Specifically, features are normalized using (\ref{x'}), with $\epsilon$ added to the denominator to avoid division by zero. Then, scaling and shifting are performed using learnable parameters $\gamma$ and $\beta$ as (\ref{y}).
Then a maximum pooling layer follows for downsampling, further reducing the computational effort.
\begin{equation}
\label{x'}
x^{\prime} = \frac{x - \mu}{\sqrt{\sigma^2 + \epsilon}},
\end{equation}
\begin{equation}
\label{y}
 y = \gamma x^{\prime} + \beta,
\end{equation}

As the scale and complexity of particle swarm increase, model accuracy tends to decline, and deepening the network architecture to improve accuracy often influence convergence effect during training. ResNet effectively addresses these gradient-related problems with residual learning and skip connections while reducing parameter count and inference time. 
We design backbone based on ResNet, which consists of four composite layers. Each layer extracts features using a varying number of BasicBlocks. The BasicBlock is a residual block structure, which contains 3*3 convolutional layers and BN layers with ReLU. The convolutional layer reduce the spatial size of feature map $f \in \mathbb{R}^{C \times H \times W}$ and highlight important features, increasing the abstraction level of features from shallow to deeper layer. BasicBlock also maintains gradient flow by constant mapping to make the features skip certain layers for propagation, accelerating convergence and improving generalization ability, and, finally, generates feature maps that are enrich with high-level semantic information and match the dimensions of downsampling layer.

We take transformer as the neck. First, 256*28*28 feature map output from backbone is spread, and then features are adjusted to match the dimension of transformer encoder through linear layers. The adjusted features are further enhanced through self-attention mechanism and feed-forward network sequences, to enable model to capture long range dependencies and improve the representation of features. After introducing transformer, the accuracy is increased by about 5\%.

Finally, we use a fully connected layer as the classification header, which is capable of converting high-dimensional features into class-specific predictions, and the output is a vector consisting of two values ranging from 0 to 1 to describe the degree of degeneracy.

\subsection{Hierarchical Anti-Degeneracy Strategies}

\subsubsection{Global Position Optimization Based on the Degree of Degeneracy}

\begin{figure}[t]\centering
	\includegraphics[width=8.8cm]{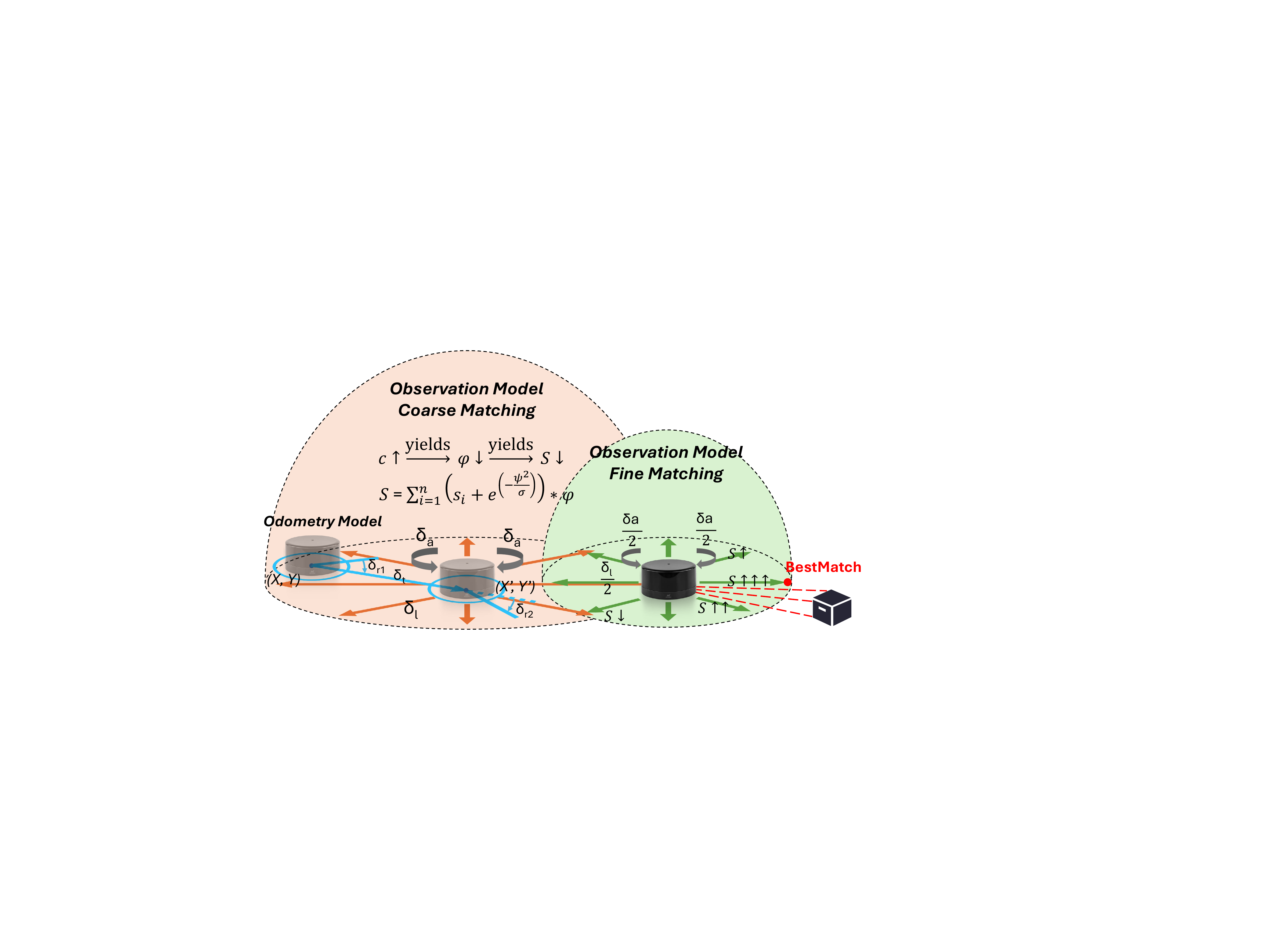}
    \vspace{-18pt}
	\caption{Schematic diagram of global position optimization.}
    \vspace{-15pt}
    \label{pose_optimize}
\end{figure}

This article proposes a global pose optimization strategy based on degeneracy confidence. The schematic is shown in Fig. \ref{pose_optimize}.
The position is first initially adjusted using the odometry model, and then the scan matching of observation model is used for further optimization.
We divide the lidar-based position optimization into a combination of coarse and fine matching. 
In the coarse matching phase, the robot will do displacements in eight directions with a step size of $\delta_l \in \mathbb{R}$ and rotations with an angle of $\delta_a \in \mathbb{S}^1$, and after each action has been performed, the current matching score $S \in \mathbb{R}^+$ will be calculated, the one with the highest score will be selected as the best position and updated, whereas in the fine matching phase, the scales of $\delta_l/2$ and $\delta_a/2$ will be used to perform the same process. 
The number of executions of the above matching combinations will adaptively change with the degeneracy level to ensure the accuracy of the search space. 
The score of scanmatch directly determines the credibility of the current pose. We scale the score of lidar through an factor $\varphi$ that includes degeneracy level $c$ in (\ref{pose_es}), so that the score of lidar is lower when the degeneracy is serious. At this time, the score of odometry pose is higher and closer to the real pose, so the odometry pose will be selected as the optimal pose. All of the above methods essentially achieve degeneracy optimization by searching for poses with higher scores, avoiding the explicit dependence on the direction of degeneracy.
\begin{equation}
\begin{cases}
\label{pose_es}
\varphi =  -e^{0.7c}+e^{0.35}+1 \\
s = \sum_{i=1}^{n} (s_i+e^{(-\psi^2 / \sigma )}) \cdot \varphi  \\
\end{cases}
\end{equation}
where $c$ is the degeneracy confidence given by the model, $s_i$ is the fraction of the $i^{th}$ beam, $\psi$ is the distance between the predicted value and the true value at the hit point, and $\mu$ is the standard deviation of the Gaussian distribution.


\subsubsection{Enhanced Resampling for Particles}


The resampling reduces particle diversity. To address this, we design an enhanced resampling method that includes selecting, fusing, and perturbing operations. First, we remove low-weight particles and keep high-weight ones. Then, for each particle, we select a particle with the closest Euclidean distance to perform a mean-crossing operation. This combines the positional information of two particles to generate new ones, increasing diversity.
Subsequently, we add a random perturbation to the fusion result to further increase the diversity, allowing the particle population to better represent the state space in which the robot may be located, as shown in (\ref{eq:random_perturbation}). 
The Gaussian form of the perturbation is more stable and balances the probability of the exploration of different regions of the state space, thus increases the probability of exploring the optimal pose.
\begin{equation}
\label{eq:random_perturbation}
\begin{cases} 
x' = x_n + \frac{1}{w } \cdot \frac{1}{\sqrt{2\pi\sigma^2}} e^{-\frac{(x-\mu)^2}{2\sigma^2}} \\
y' = y_n + \frac{1}{w } \cdot \frac{1}{\sqrt{2\pi\sigma^2}} e^{-\frac{(y-\mu)^2}{2\sigma^2}} \\
\end{cases}
\end{equation}
where $(x_n,y_n) \in \mathbb{R}^2$ are the coordinates after performing mean crossing, $(x',y') \in \mathbb{R}^2$ are the new coordinates after adding the perturbation, $w$ represents the particle weights.

\begin{table*}[]
\centering
\caption{Results of the ablation experiments.}
\vspace{-8pt}
\label{performance_comparison}
\resizebox{\textwidth}{!}{%
\begin{tabular}{cccccccccccc}
\hline
 &
   &
   &
   &
   &
  \multicolumn{2}{c}{Loss} &
  \multicolumn{2}{c}{Optimizer} &
   &
   &
   \\ \cline{6-7} \cline{8-9}
\multirow{-2}{*}{Model} &
  \multirow{-2}{*}{\begin{tabular}[c]{@{}c@{}}Linear\\ Mapping\end{tabular}} &
  \multirow{-2}{*}{\begin{tabular}[c]{@{}c@{}}Gaussian \\ Augmentation\end{tabular}} &
  \multirow{-2}{*}{\begin{tabular}[c]{@{}c@{}}GPU\\ Acceleration\end{tabular}} &
  \multirow{-2}{*}{\begin{tabular}[c]{@{}c@{}}Image\\ Matrix\end{tabular}} &
  \multicolumn{1}{c}{CrossEntropyLoss} &
  BCELoss &
  \multicolumn{1}{c}{Adam} &
  RMSprop &
  \multirow{-2}{*}{Accuracy} &
  \multirow{-2}{*}{F1 Score} &
  \multirow{-2}{*}{Time(ms)} \\ \hline
 &
  \cellcolor[HTML]{FFCCC9}$\checkmark$ &
  \cellcolor[HTML]{FFCCC9}$\checkmark$ &
  \cellcolor[HTML]{CBCEFB}- &
  \cellcolor[HTML]{CBCEFB}- &
  \multicolumn{1}{c}{$\checkmark$} &
  - &
  \multicolumn{1}{c}{$\checkmark$} &
  - &
  0.901 &
  0.851 &
  20 \\ 
 &
  \cellcolor[HTML]{FFCCC9}$\checkmark$ &
  \cellcolor[HTML]{FFCCC9}$\checkmark$ &
  \cellcolor[HTML]{CBCEFB}- &
  \cellcolor[HTML]{CBCEFB}- &
  \multicolumn{1}{c}{$\checkmark$} &
  - &
  \multicolumn{1}{c}{-} &
  $\checkmark$ &
  0.8965 &
  0.839 &
  24 \\ 
 &
  \cellcolor[HTML]{FFCCC9}$\checkmark$ &
  \cellcolor[HTML]{FFCCC9}$\checkmark$ &
  \cellcolor[HTML]{CBCEFB}- &
  \cellcolor[HTML]{CBCEFB}- &
  \multicolumn{1}{c}{-} &
  $\checkmark$ &
  \multicolumn{1}{c}{$\checkmark$} &
  - &
  0.8934 &
  0.853 &
  25 \\ 
\multirow{-4}{*}{MobileNetV3} &
  \cellcolor[HTML]{FFCCC9}$\checkmark$ &
  \cellcolor[HTML]{FFCCC9}$\checkmark$ &
  \cellcolor[HTML]{CBCEFB}- &
  \cellcolor[HTML]{CBCEFB}- &
  \multicolumn{1}{c}{-} &
  $\checkmark$ &
  \multicolumn{1}{c}{-} &
  $\checkmark$ &
  0.8828 &
  0.85 &
  22 \\ \hline
 &
  \cellcolor[HTML]{FFCCC9}$\checkmark$ &
  \cellcolor[HTML]{FFCCC9}$\checkmark$ &
  \cellcolor[HTML]{CBCEFB}- &
  \cellcolor[HTML]{CBCEFB}- &
  \multicolumn{1}{c}{$\checkmark$} &
  - &
  \multicolumn{1}{c}{$\checkmark$} &
  - &
  - &
  - &
  130 \\ 
 &
  \cellcolor[HTML]{FFCCC9}$\checkmark$ &
  \cellcolor[HTML]{FFCCC9}$\checkmark$ &
  \cellcolor[HTML]{CBCEFB}- &
  \cellcolor[HTML]{FFCCC9}$\checkmark$ &
  \multicolumn{1}{c}{$\checkmark$} &
  - &
  \multicolumn{1}{c}{$\checkmark$} &
  - &
  - &
  - &
  45 \\ 
 &
  \cellcolor[HTML]{FFCCC9}$\checkmark$ &
  \cellcolor[HTML]{FFCCC9}$\checkmark$ &
  \cellcolor[HTML]{FFCCC9}$\checkmark$ &
  \cellcolor[HTML]{CBCEFB}- &
  \multicolumn{1}{c}{$\checkmark$} &
  - &
  \multicolumn{1}{c}{$\checkmark$} &
  - &
  - &
  - &
  112 \\ 
 &
  \cellcolor[HTML]{FFCCC9}$\checkmark$ &
  \cellcolor[HTML]{FFCCC9}$\checkmark$ &
  \cellcolor[HTML]{FFCCC9}$\checkmark$ &
  \cellcolor[HTML]{FFCCC9}$\checkmark$ &
  \multicolumn{1}{c}{$\checkmark$} &
  - &
  \multicolumn{1}{c}{-} &
  $\checkmark$ &
  0.975 &
  0.933 &
  {\color[HTML]{3531FF} \textbf{14}} \\ 
 &
  \cellcolor[HTML]{FFCCC9}$\checkmark$ &
  \cellcolor[HTML]{FFCCC9}$\checkmark$ &
  \cellcolor[HTML]{FFCCC9}$\checkmark$ &
  \cellcolor[HTML]{FFCCC9}$\checkmark$ &
  \multicolumn{1}{c}{-} &
  $\checkmark$ &
  \multicolumn{1}{c}{$\checkmark$} &
  - &
  0.979 &
  0.942 &
  15 \\ 
 &
  \cellcolor[HTML]{FFCCC9}$\checkmark$ &
  \cellcolor[HTML]{FFCCC9}$\checkmark$ &
  \cellcolor[HTML]{FFCCC9}$\checkmark$ &
  \cellcolor[HTML]{FFCCC9}$\checkmark$ &
  \multicolumn{1}{c}{-} &
  $\checkmark$ &
  \multicolumn{1}{c}{-} &
  $\checkmark$ &
  {\color[HTML]{3531FF} \textbf{0.981}} &
  {\color[HTML]{3531FF} \textbf{0.963}} &
  17 \\ 
 &
  \cellcolor[HTML]{FFCCC9}$\checkmark$ &
  \cellcolor[HTML]{FFCCC9}$\checkmark$ &
  \cellcolor[HTML]{FFCCC9}$\checkmark$ &
  \cellcolor[HTML]{FFCCC9}$\checkmark$ &
  \multicolumn{1}{c}{$\checkmark$} &
  - &
  \multicolumn{1}{c}{$\checkmark$} &
  - &
  {\color[HTML]{FE0000} \textbf{0.993}} &
  {\color[HTML]{FE0000} \textbf{0.972}} &
  {\color[HTML]{FE0000} \textbf{10}} \\ 
 &
  \cellcolor[HTML]{CBCEFB}- &
  \cellcolor[HTML]{CBCEFB}- &
  \cellcolor[HTML]{FFCCC9}$\checkmark$ &
  \cellcolor[HTML]{FFCCC9}$\checkmark$ &
  \multicolumn{1}{c}{$\checkmark$} &
  - &
  \multicolumn{1}{c}{$\checkmark$} &
  - &
  0.826 &
  0.783 &
  - \\ 
 &
  \cellcolor[HTML]{CBCEFB}- &
  \cellcolor[HTML]{FFCCC9}$\checkmark$ &
  \cellcolor[HTML]{FFCCC9}$\checkmark$ &
  \cellcolor[HTML]{FFCCC9}$\checkmark$ &
  \multicolumn{1}{c}{$\checkmark$} &
  - &
  \multicolumn{1}{c}{$\checkmark$} &
  - &
  0.852 &
  0.823 &
  - \\ 
\multirow{-10}{*}{DD-Model(Ours)} &
  \cellcolor[HTML]{FFCCC9}$\checkmark$ &
  \cellcolor[HTML]{CBCEFB}- &
  \cellcolor[HTML]{FFCCC9}$\checkmark$ &
  \cellcolor[HTML]{FFCCC9}$\checkmark$ &
  \multicolumn{1}{c}{$\checkmark$} &
  - &
  \multicolumn{1}{c}{$\checkmark$} &
  - &
  0.912 &
  0.89 &
  - \\ \hline
\end{tabular}%
}
\vspace{-5pt}
\end{table*}

\section{Experimental studies and results}


We use a laptop with Intel i7-7700HQ CPU and NVIDIA GeForce GTX1060 GPU to remotely maneuver an wheeled robot for data collection and validation. 
The robot is equipped with LS-M10P lidar, ARM Cortex-A72 64-bit CPU and Broadcom VideaCore VI GPU, as shown in Fig. \ref{dataset&instruments}.
In simulations, we adopt a computer with 13th Gen Intel Core i7-13700K CPU and NVIDIA GeForce RTX2060 super GPU.

\subsection{Performance of Degeneracy Detection}

\subsubsection{Dataset Generation}

\begin{figure}[t]\centering

	\includegraphics[width=8.8cm]{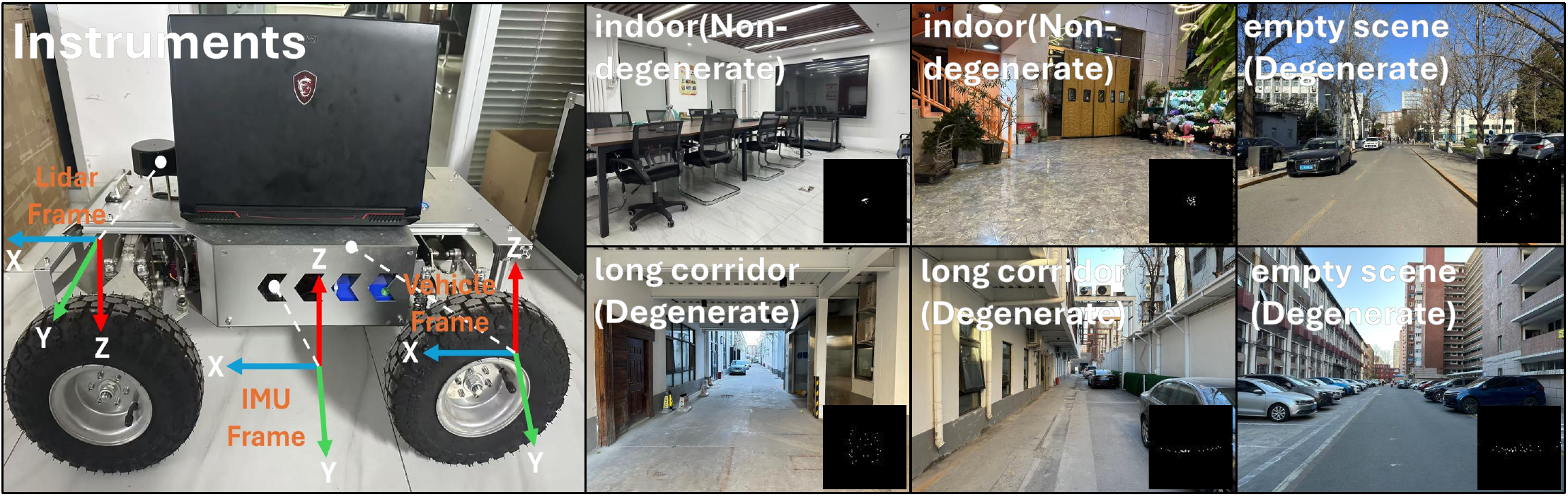}
    
	\caption{Examples of data collection scenarios and datasets.}\vspace{-17pt}\label{dataset&instruments}
\end{figure}

We record the coordinates of particle in different scenes, as annotations shown in Fig. \ref{dataset&instruments}, and label them with binary values according to $R$ in (\ref{label_explain}). Since particle weight $w$ reflects credibility, a larger variance and smaller mean of $w$ suggest higher degeneracy. We find $R$ is usually greater than 1 in degenerate scenes, so we classify data with $R>1$ as degenerate and data with $R<1$ as non-degenerate. Finally, there are 3937 degenerate images and 3827 non-degenerate images, the ratio of training, test and validation set is 6:2:2. 
\begin{equation}
\label{label_explain}
R = \frac { \frac { 1 } { N } \sum _ { i = 1 } ^ { N } ( w _ { i } - \overline { w } ) ^ { 2 } } { \frac { 1 } { N } \sum _ { i = 1 } ^ { N } w _ { i } }
\end{equation}
where $\overline { w }$ is the mean of $w$ and $N$ is particle number.

\begin{figure*}[htb]\centering
	\includegraphics[width=18cm]{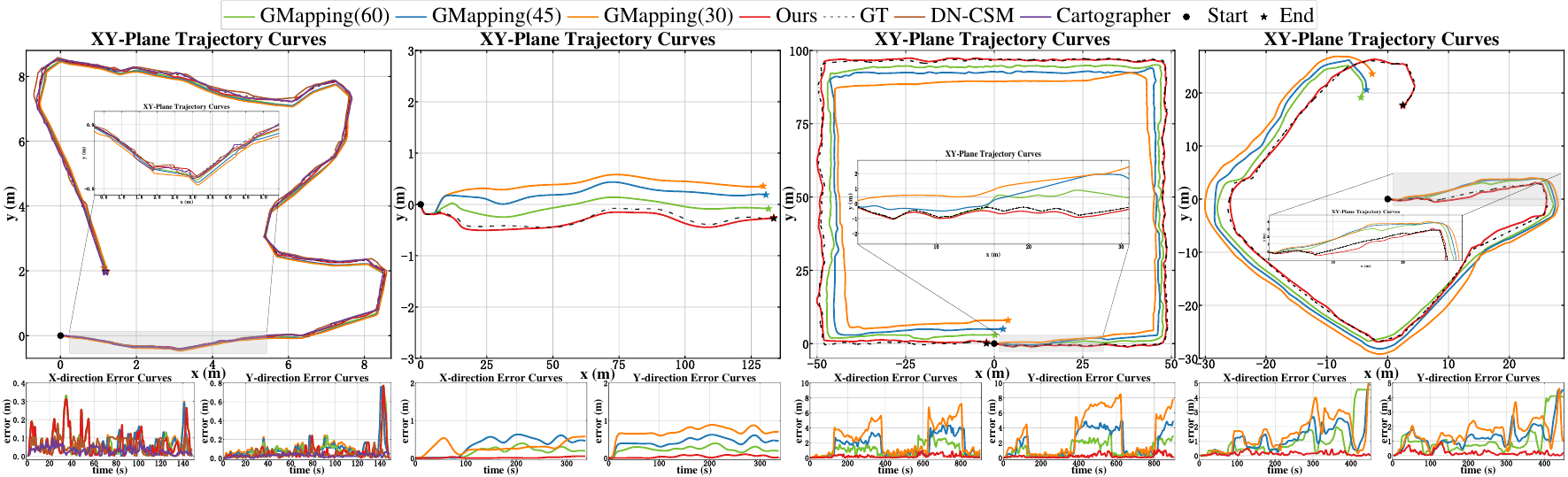}
    \vspace{-7pt}
	\caption{Evaluation plot of trajectory accuracy for simulation experiments.}\label{simul_locate_result}
\end{figure*}

\begin{table*}[htb]
\centering
\fontsize{7.5}{5}\selectfont
\begin{threeparttable}
\vspace{-10pt}
\caption{Positioning errors in simulation experiments.}
\label{simul_com}
\begin{tabular*}{\textwidth}{@{\extracolsep{\fill}}cccccccccccccc}
	\toprule
 
	\multirow{3}{*}{\textbf{Method}}
    
    &\multicolumn{3}{c}{\textbf{Indoor environment}}
    &\multicolumn{3}{c}{\textbf{Straight corridor}}
    &\multicolumn{3}{c}{\textbf{Circular corridor}}
    &\multicolumn{3}{c}{\textbf{Outdoor open environment}}\cr

	\cmidrule(l){2-4} 
	\cmidrule(l){5-7} 
	\cmidrule(l){8-10} 
	\cmidrule(l){8-10} 
	\cmidrule(l){11-13} 
    
	&ATE&$Error_x$&$Error_y$&ATE&$Error_x$&$Error_y$&ATE&$Error_x$&$Error_y$&ATE&$Error_x$&$Error_y$\cr

	\midrule
	\multirow{1}{*}{GMapping(60)}
    
    &0.1319&0.07183&0.08162
    &0.22059&0.17182&0.3505
    &1.99140&0.98137&1.063
    &2.78390&1.05844&1.10095\cr
    
	\multirow{1}{*}{GMapping(45)}
    
    &0.1373&0.07405&0.08959
    &0.47990&0.25928&0.45251
    &3.61930&1.60959&1.92411
    &3.27640&1.15916&1.42285\cr
    
	\multirow{1}{*}{GMapping(30)}
    
    &0.1498&0.0780&0.1042
    &0.75860&0.30329&0.7886
    &5.68730&2.40482&3.03794
    &3.54170&1.49342&1.58323\cr

    \multirow{1}{*}{Cartographer}
    
    &0.106&0.0323&0.0275
    &-&-&-
    &-&-&-
    &-&-&-\cr

    \multirow{1}{*}{DN-CSM}
    
    &0.1276&0.0496&0.0383
    &-&-&-
    &-&-&-
    &-&-&-\cr

    \multirow{1}{*}{Ours}
    
    &0.1180&0.0384&0.0402
    &0.15760&0.10483&0.11407
    &0.55660&0.18372&0.20154
    &0.72270&0.14142&0.19495\cr
    
	\bottomrule
\end{tabular*}
\end{threeparttable}
\vspace{-15pt}
\end{table*}

\begin{table}[]
\centering
\caption{Comparison of Switch-SLAM and ours.}
\vspace{-7pt}
\label{detection_comparsion}
\resizebox{\columnwidth}{!}{%
\begin{tabular}{ccccc}
\hline
                     & Accuracy & Precision & Recall & F-1 Score \\ \hline
Switch-SLAM {\cite{lee2024switch}} & 0.775    & 0.767     & 0.735  & 0.751    \\ 
Ours                 & \textbf{0.965}    & \textbf{0.971}     & \textbf{0.945}  & \textbf{0.952}     \\ \hline
\end{tabular}%
}
\vspace{-12pt}
\end{table}

\subsubsection{Ablation Experiments}

We evaluate the contribution of different components in DD-Model by ablation experiments. We test various loss functions and optimizers on two architectures: MobileNetV3 and DD-Model. Accuracy, F1 score and inference time on validation set are shown in Table. \ref{performance_comparison}. Specifically, we use MobileNetV3\_Large and adapt it to degeneracy detection task. 
The results show that MobileNetV3 achieve about 90\% accuracy and 0.85 F1 score, showing good performance but with room for improvement. In contrast, DD-Model achieve 99.3\% accuracy and 0.972 F1 score, demonstrating superior detection capability and ability to identify vast majority of degenerate states. Based on these results, we select CrossEntropyLoss and Adam as hyperparameters.

We also test the impact of GPU acceleration and image matrix. The results show these methods reduce inference time from 130ms to 112ms and 45ms respectively, with the fastest speed reaching 10ms, ensuring the real-time performance. 
Finally, experiments show that directly feeding the coordinates to model without linear mapping or removing Gaussian augmentation reduces accuracy by about 14.1\% and 8.1\%, respectively. This demonstrates the contribution of linear mapping and Gaussian augmentation to model accuracy.

\begin{figure}[t]\centering
	\includegraphics[width=8.8cm]{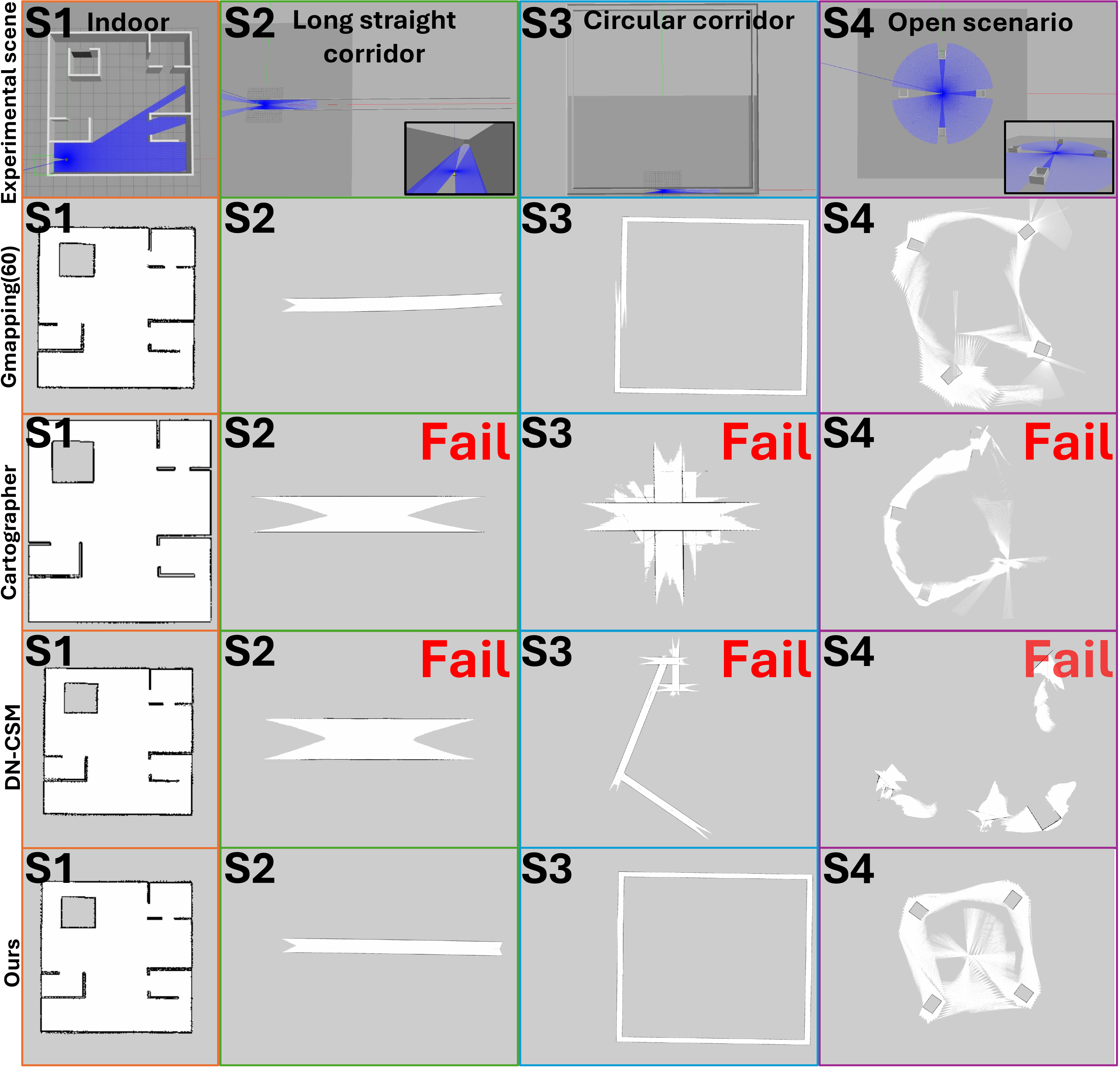}
    \vspace{-18pt}
	\caption{The simulation scenarios and the constructed maps.}
    \vspace{-18pt}
    \label{simul_ex}
\end{figure}

\begin{figure*}[htb]\centering
	\includegraphics[width=18cm]{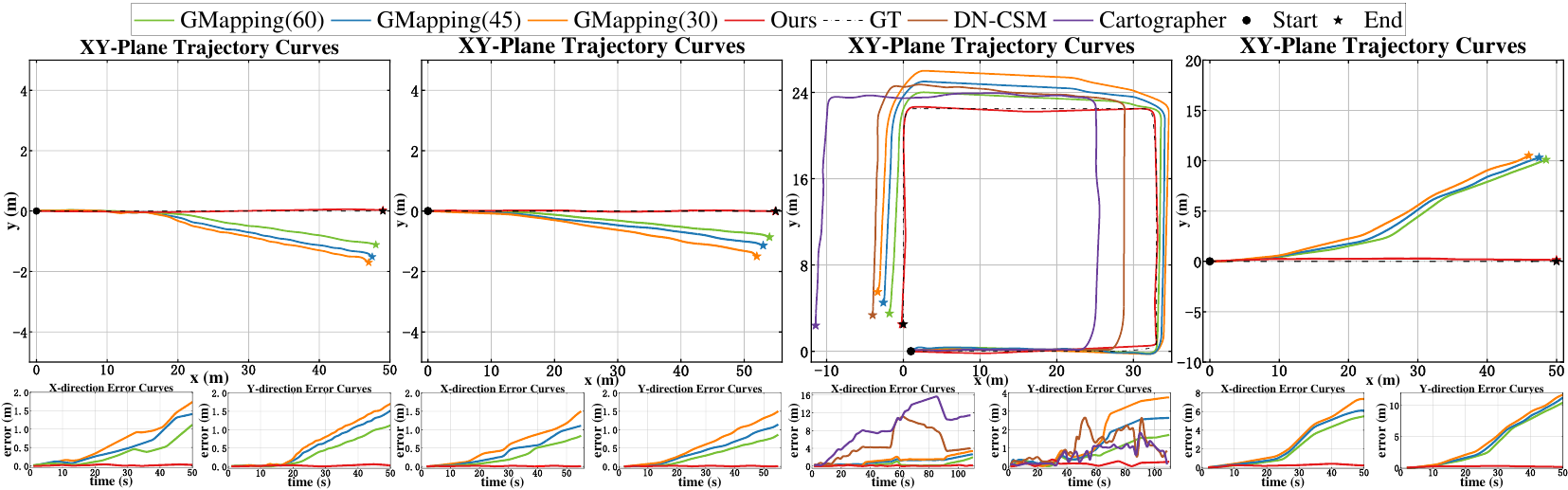}
    \vspace{-8pt}
	\caption{Evaluation plot of trajectory accuracy for actual experiments.}
    \vspace{-10pt}
    \label{actual_locate_result}
\end{figure*}

\begin{table*}[htb]
	
\centering
\fontsize{7.5}{5}\selectfont
\begin{threeparttable}
\caption{Positioning errors in actual experiments.}
\label{actual_com}
\begin{tabular*}{\textwidth}{@{\extracolsep{\fill}}cccccccccccccc}
	\toprule
 
	\multirow{3}{*}{\textbf{Method}}
    
    &\multicolumn{3}{c}{\textbf{Straight corridor A}}
    &\multicolumn{3}{c}{\textbf{Straight corridor B}}
    &\multicolumn{3}{c}{\textbf{Circular corridor}}
    &\multicolumn{3}{c}{\textbf{Outdoor open environment}}\cr

	\cmidrule(l){2-4} 
	\cmidrule(l){5-7} 
	\cmidrule(l){8-10} 
	\cmidrule(l){8-10} 
	\cmidrule(l){11-13} 
    
	&ATE&$Error_x$&$Error_y$&ATE&$Error_x$&$Error_y$&ATE&$Error_x$&$Error_y$&ATE&$Error_x$&$Error_y$\cr

	\midrule
	\multirow{1}{*}{GMapping(60)}
    
    &0.5481&0.3146&0.39707
    &0.6806&0.33272&0.41618
    &0.8391&0.42145&0.75231
    &5.2080&2.12375&3.91135\cr
    
	\multirow{1}{*}{GMapping(45)}
    
    &0.7613&0.50234&0.56628
    &0.7819&0.41485&0.54816
    &1.4794&1.0563&1.22102
    &5.5455&2.51211&4.24782\cr
    
	\multirow{1}{*}{GMapping(30)}
    
    &0.8864&0.6478&0.6782
    &0.9285&0.5699&0.66769
    &2.0767&1.27385&1.64918
    &5.9426&2.86529&4.65336\cr

    \multirow{1}{*}{Cartographer}
    
    &-&-&-
    &-&-&-
    &9.215&8.7964&0.6779
    &-&-&-\cr

    \multirow{1}{*}{DN-CSM}
    
    &-&-&-
    &-&-&-
    &5.7387&5.0225&1.0977
    &-&-&-\cr

    \multirow{1}{*}{Ours}
    
    &0.2313&0.01996&0.02598
    &0.5520&0.01237&0.01297
    &0.4683&0.08895&0.1624
    &0.8229&0.22784&0.20634\cr
    
	\bottomrule
\end{tabular*}
\end{threeparttable}
\vspace{-17pt}
\end{table*}

\subsubsection{Accuracy of Degeneracy Detection}

In degeneracy detection experiment, we use \cite{lee2024switch} as baseline and replace the eigenvalues of Hessian matrix with those of the covariance matrix $\Sigma$ of particle weights in (\ref{w_eigen}). This is because these eigenvalues can reflect the uncertainty in particle estimation.
\begin{equation}
\label{w_eigen}
\Sigma = \frac { 1 } { N - 1 } \sum _ { i = 1 } ^ { N } ( w _ { i } - \overline { w } ) ( w _ { i } - \overline { w } ) ^ { T }
\end{equation}
where $w_i$ is the weight of $i_{th}$ particle, $\Sigma$ is the covariance matrix of weights and $\overline { w }$ is the mean of weights.

The comparative results on validation set are shown in Table. \ref{detection_comparsion}. 
Our method is better to \cite{lee2024switch} in all metrics, and the accuracy is increased by 19\%, which proves the superior detection capability of DD-Model.

\begin{figure}[!t]\centering

	\includegraphics[width=8.8cm]{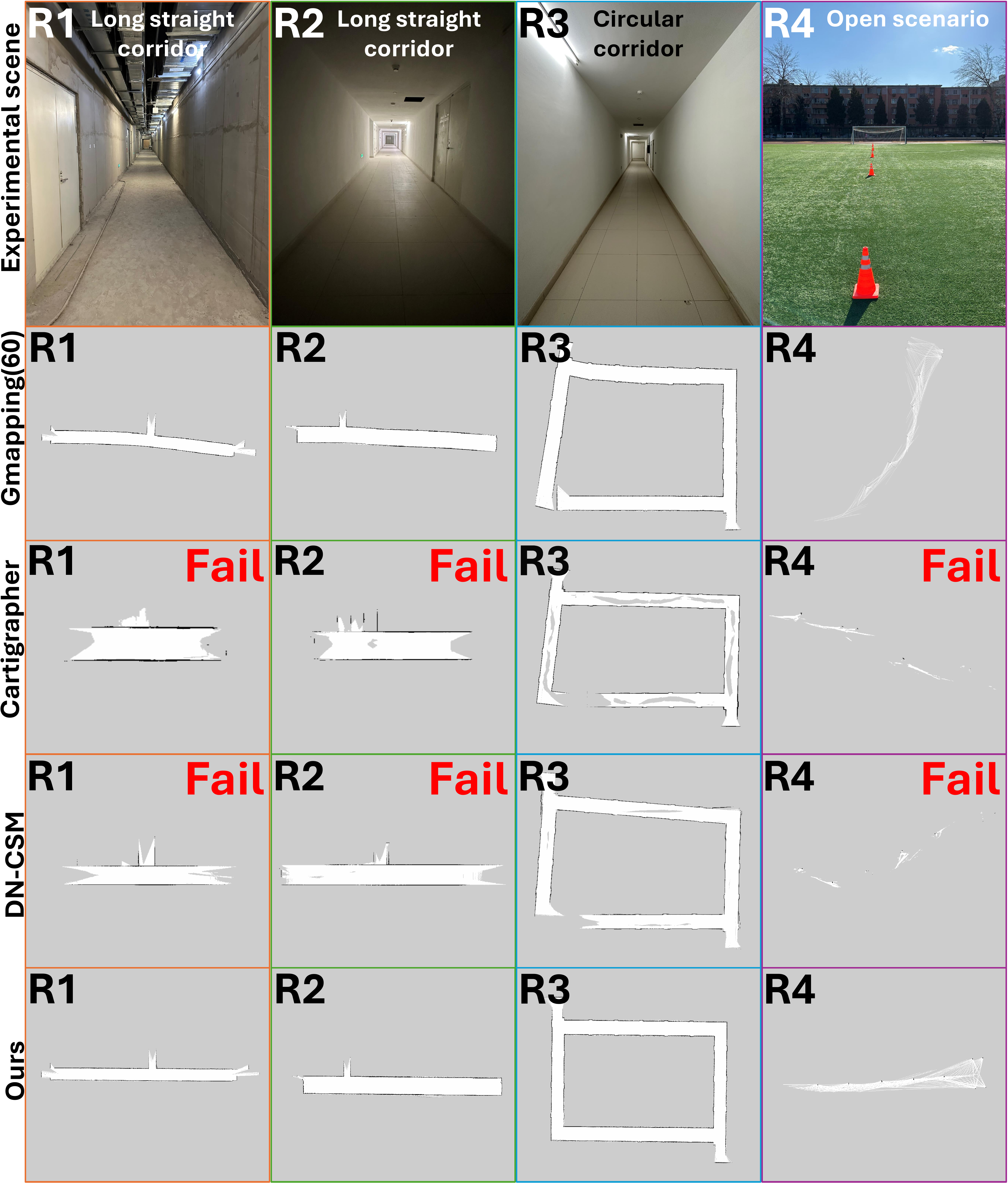}
    \vspace{-18pt}
	\caption{The real-world scenarios and the constructed maps.}\vspace{-23pt}\label{actual_ex}
\end{figure}

\subsection{Performance of Anti-Degeneracy Systems in Simulation Environments}

We set up four simulation environments S1-S4 to carry out the degeneracy optimization experiments.
As shown in Fig. \ref{simul_ex}, S1 is 10m*10m indoor environment, S2 is 140m*1m long straight corridor, S3 is 100m*100m circular corridor and S4 is open scenario with sparse landmarks.
We conduct comparative experiments with GMapping \cite{grisetti2007improved}, cartographer \cite{7487258} and DN-CSM \cite{shi2022dense}, while our method uses 30 particles. We plot the maps given by GMapping with 60 particles and other methods in Fig. \ref{simul_ex}, and also plot the localization trajectories of all methods and the ground truth given by gazebo in Fig. \ref{simul_locate_result}, including GMapping with 30, 45 and 60 particles. 
In addition, the absolute value of the difference between each trajectory and true value in x and y dimensions is plotted in Fig. \ref{simul_locate_result}, and its mean value ($Error_x$, $Error_y$) is shown in Table. \ref{simul_com}.
Finally, the accuracy is quantitatively evaluated by absolute trajectory error (ATE) using evo \cite{grupp2017evo}. Specifically, since position accuracy is the primary concern in robot navigation tasks, we only consider the translational component of trajectory in ATE.

It can be seen that all methods have similar performance in S1 and can accurately construct indoor maps. Cartographer, which employs loop closure detection, has the smallest ATE of 0.106. 
In S2-S4, both cartographer and DN-CSM fail, unable to construct the basic environmental structure, with the map showing varying degrees of stagnation and steering errors in the direction of movement. 
In S2, the map constructed by GMapping(60) is skewed. Compared to GMapping(30), our method reduces ATE by up to 79.2\%, with $Error_x$ and $Error_y$ maintained around 0.1, enabling the construction of accurate map. 
In S3, loop closure in the map by GMapping(60) is not closed, whereas our method constructs accurate loop structure, achieving ATE reduction of up to 90.2\% and the best fit with GT trajectory. 
In S4, our method has low $Error_x$ and $Error_y$, with localization trajectory closest to GT, and ATE reduction of 79.6\% compared to GMapping(30). 
Additionally, we record the average inference time of DD-Model and hierarchical anti-degeneracy strategies, which are 15.6ms and 25.1ms respectively. 
The above experiments demonstrate that our method has high localization accuracy in various scenarios and can mitigate the negative impact of degenerate environments on SLAM while ensuring real-time performance.

\subsection{Performance of Anti-Degeneracy Systems in Real-World Environments}

We also conduct degeneracy optimization experiments in four real-world environments R1-R4 shown in Fig. \ref{actual_ex}, R1 is 50m*2.5m long corridor, R2 is 60m*3m long corridor, R3 is 34m*24m circular corridor and R4 is outdoor scenario with sparse landmarks. In terms of evaluation metrics, it is the same as simulations, and GT of trajectory is manually measured.

As shown in Fig. \ref{actual_ex}, in R1 and R2, cartographer and DN-CSM show movement stagnation and lose localization in the movement direction. Similarly, in R4, they fail to build correct positions between landmarks because of localization drift.

In R1, GMapping(60) has no localization stagnation, but it shows misalignment due to sparse features, resulting in a skewed map. On the contrary, our method build accurate straight corridor structure. As shown in Table. \ref{actual_com}, compared with GMapping(30), ATE of our method decreases by about 73.9\%, $Error_x$ and $Error_y$ are lower than 0.03. 
In R2, GMapping(60) also shows localization deviation. As can be seen from Fig. \ref{actual_locate_result}, the error between localization trajectory and GT of our method is the smallest, the error levels of $Error_x$ and $Error_y$ are the lowest, and ATE is reduced by 40.5\% at the highest. 
In R3, map loops constructed by GMapping(60), cartographer, and DN-CSM all show dislocations to varying degrees. As can be seen from Table. \ref{actual_com}, ATE of cartographer is the highest, 9.215, followed by DN-CSM, 5.7387. Compared with these two methods, the ATE of our method is reduced by 94.9\% and 91.8\% respectively, and finally more accurate localization results and loop maps are obtained. 
In R4, the relative position of landmarks constructed by GMapping(60) is wrong. Our method not only shows the correct location of landmarks, but also reduces the ATE by about 86.2\% compared with GMapping(30) to obtain more accurate localization results. 
In addition, we record the mean inference time of DD-Model and hierarchical anti-degeneracy strategies, which are 19.3ms and 27.2ms, respectively.

The above experiments show that compared with GMapping, cartographer, and DN-CSM, our method can improve the localization accuracy of SLAM in various environments and ensure real-time performance in real world. Meanwhile, its adaptive features overcome the heavy dependence of cartographer on preset parameters and the dependence of GMapping on the number of particles.

\section{Conclusion}

This article has presented a deep learning-based anti-degeneracy SLAM. We first proposed linear mapping and Gaussian-based data augmentation to ensure high-quality dataset. Next, we designed DD-Model using ResNet and transformer, which effectively classifies particle distributions to determine degeneracy levels. Finally, we designed a hierarchical anti-degeneracy strategy that enhances resampling, providing accurate initial values for pose optimization. The optimization process can adaptively adjust the frequency and sensor trustworthiness based on degeneracy levels.
We demonstrated the optimality of DD-Model and the contribution of each module through ablation experiments. In addition, the results of simulation and real-world experiments showed that the anti-degeneracy SLAM is able to improve global pose estimation and reduce localization drift in various environments. In the future, we will improve the generalization ability of model under conditions of extreme particle numbers.






\bibliographystyle{Bibliography/IEEEtranTIE}
\bibliography{Bibliography/IEEEabrv,references.bib}\ 

\end{document}